# Hippocampal Atrophy Patterns Across the Alzheimer's Disease Spectrum: A Voxel-Based Morphometry Analysis

*Trishna Niraula*

Arkansas State University, Jonesboro, AR, USA

**ABSTRACT**

Alzheimer's disease (AD) and mild cognitive impairment (MCI) are associated with progressive gray matter loss, particularly in medial temporal structures. In this study, CAT12/SPM12 voxel-based morphometry was applied to baseline T1-weighted MRI scans from 249 ADNI participants (CN = 90, MCI = 129, AD = 30). Gray matter volume was analyzed using a general linear model, with the diagnostic group as primary predictor and age and total intracranial volume as covariates. Statistical maps were thresholded at $p < 0.001$ (voxelwise) and corrected for multiple comparisons at the cluster level using family-wise error (FWE) correction ($p < 0.05$). Significant hippocampal atrophy was observed in AD relative to CN and MCI (Cohen's d = 2.03 and 1.61, respectively). Hippocampal volume demonstrated moderate predictive value for conversion from MCI to AD (AUC = 0.66). Stratification by APOE4 status did not reveal significant genetic effects on cross-sectional hippocampal volume. These results support medial temporal degeneration as a key feature of AD progression and provide insights into predictive biomarkers and genetic influences.

*Index Terms*— Voxel Based Morphometry (VBM), Computational Anatomy Toolbox, Alzheimer's Disease (AD), MCI, CN

## 1. INTRODUCTION

Alzheimer's disease (AD) is the most common cause of dementia, and is characterized by progressive memory loss and cognitive decline [1], affecting over 6 million Americans [12]. The disease involves pathological hallmarks including amyloid-β plaques, neurofibrillary tangles, and progressive neuronal loss [13,14]. Early stages of the disease often appear as Mild Cognitive Impairment (MCI), which represents a transitional phase between normal aging and Alzheimer's. Structural brain imaging provides a valuable approach to study these changes, and voxel-based morphometry (VBM) is a popular method for assessing group differences in brain structure, especially in gray matter volume (GMV) [2,10].

Early VBM studies in Alzheimer's disease mainly looked at the medial temporal regions, such as the hippocampus and entorhinal cortex, because of their well-known role in memory decline. However, recent reviews have also highlighted the involvement of the prefrontal cortex. One systematic review found that about 76% of studies reported gray matter loss in the frontal lobe [6], including areas like the inferior, middle, and superior frontal gyri, the anterior cingulate cortex, and the medial and orbitofrontal cortices. Some of these frontal changes can even be seen at the MCI stage, before a full Alzheimer's diagnosis. Frontal atrophy has been linked to declines in executive function, cognition, and neuropsychiatric symptoms [5,6], showing that Alzheimer's-related brain changes affect more than just the temporolimbic areas and involve broader cortical networks.

Despite the expansion of VBM studies in AD/MCI, several persistent limitations have been noted in the literature. Many earlier studies used small or mixed clinical groups, did not confirm diagnoses with biomarkers, and relied on cross-sectional designs, which makes it hard to study disease progression. Differences in preprocessing and statistical methods have also made it difficult to reproduce results and compare findings across studies. Researchers have suggested using larger datasets with biomarker validation, conducting longitudinal studies to track changes over time, and adopting consistent imaging and statistical methods to make results more reliable and useful [5,7].

## 2. METHODS

This study utilized structural magnetic resonance imaging (MRI) and voxel-based morphometry (VBM) using MATLAB R2025 based Computational Anatomy Toolbox (CAT12) for Statistical Parametric Mapping (SPM) to identify and quantify regional gray matter volume (GMV) differences between cognitively normal (CN), mild cognitive impairment (MCI), and Alzheimer's Disease (AD) groups from the ADNI cohort [8,9]. VBM allows for an unbiased, whole-brain assessment of morphological differences across subject groups by comparing the local concentration of gray matter [15]. After bias-field correction and unified adaptive segmentation into gray matter (GM), white matter (WM), and cerebrospinal fluid (CSF), images were spatially normalized to MNI space using fast geodesic shooting. Modulation with Jacobian determinants preserved native tissue volumes, and the resulting GMV maps were smoothed with an 8 mm FWHM Gaussian kernel. Quality control procedures, including quantitative metrics and visual inspection, excluded subjects with corrupted or incomplete 3D volumes. Voxel-wise statistical analyses were then conducted via two-sample t-tests to produce maps of significant atrophy, significant clusters were delineated as regions of interest and their adjusted eigenvariates (summary voxel intensity representations) were extracted for further group comparisons.

## 2.1 DATA COLLECTION

Data used in the preparation of this article were obtained from the Alzheimer's Disease Neuroimaging Initiative (ADNI) database (adni.loni.usc.edu). The ADNI was launched in 2003 as a public-private partnership, led by Principal Investigator Michael W. Weiner, MD [3]. The ADNI study aimed to test whether serial magnetic resonance imaging (MRI), positron emission tomography (PET), other biological markers, and clinical and neuropsychological assessments can be combined to measure the progression of mild cognitive impairment (MCI) and early Alzheimer's disease (AD). Participants in ADNI underwent comprehensive baseline evaluations including structural MRI, cognitive testing, and biomarker collection. The ADNI protocol has been approved by institutional review boards at all participating sites, and all participants provided written informed consent.

After quality control and exclusion of subjects due to non-3D acquisitions, preprocessing failure, and missing metadata, the final sample included 249 subjects: 30 with clinically diagnosed Alzheimer's disease (AD), 129 with mild cognitive impairment (MCI), and 90 cognitively normal (CN) individuals. All selected images were acquired on 3T MRI scanners, and only baseline scans were used. Participants were age and gender-matched as closely as possible to control for demographic confounds.

## 2.2 MRI PREPROCESSING

Structural MR images were processed using the Computational Anatomy Toolbox (CAT12) integrated with Statistical Parametric Mapping (SPM12) in MATLAB. The preprocessing steps included bias field correction, segmentation into gray matter (GM), white matter (WM), and cerebrospinal fluid (CSF), spatial normalization to MNI space via fast geodesic shooting.

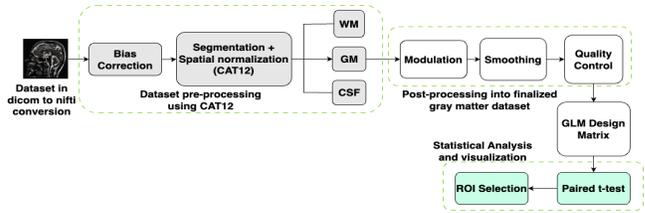

**Fig. 1.** CAT-12 work-flow

Modulation (Jacobian correction) preserved native tissue volumes, and the resulting gray matter volume maps were smoothed with an 8 mm full-width at half maximum (FWHM) Gaussian kernel. Visual and quantitative quality control excluded problematic scans.

Automated quality assessment was performed on each preprocessed scan using CAT12's built-in metrics, including image quality rating (IQR) and noise-contrast ratio. Visual inspection was conducted to identify artifacts such as motion blur, incomplete brain coverage, or registration failures. Scans with corrupted or non-3D DICOM files were excluded, along with those exhibiting preprocessing failures indicated by CAT12 error flags, excessive motion artifacts, or missing demographic metadata required for statistical modeling.

## 2.3 STATISTICAL MODELING

Statistical analysis was performed using the General Linear Model (GLM) in SPM12 [11]. The group was treated as a categorical variable with CN as the implicit reference. Covariates such as age and total intracranial volume (TIV) were included to control for individual differences.

Thus the model at each voxel was:

$$Y = X\beta + \varepsilon,$$

Where,

- Y is an n x 1 vector of modulated gray matter intensities across subjects,
- X is the n x p design matrix,
- $\beta$ is a p x 1 vector of regression coefficients, and
- $\varepsilon \sim N(0, \sigma^2 I)$ is voxel-wise Gaussian noise.

*2.3.1. Design Matrix*

The design matrix (Figure 2, Figure 4) had five columns:

**Table 1.** Design-Matrix

| Column | Name | Coding |
|--------|------|--------|
| 1 | Intercept | Baseline, CN |

| 2 | MCI Indicator | 1 for MCI, 0 otherwise |
| 3 | AD Indicator | 1 for AD, 0 otherwise |
| 4 | Age | Mean-centered age |
| 5 | TIV | Mean-centered total intracranial volume |

*2.3.2. Contrast Definitions*

Pairwise group comparisons were constructed as shown in Table 2.

*2.3.3. Thresholding and localization*

Statistical maps were thresholded at voxelwise p < 0.001 uncorrected, and inference was based on cluster-level family-wise-error (FWE) correction at p < 0.05 using Gaussian Random Field theory.

**Table 2.** Contrast definition and hypothesis, [Group1, Group2, Age, TIV]

| Contrast (Group1 > Group2) | Vector | Hypothesis |
|---|---|---|
| CN>AD | [1,-1,0,0] | CN has greater GMV than AD |
| MCI>AD | [1,-1,0,0] | MCI has greater GMV than AD |
| CN>MCI | [1,-1,0,0] | CN has greater GMV than MCI |

No additional arbitrary cluster extent threshold was imposed beyond that implicit in the estimated smoothness. Regions were localized using the Neuromorphometrics atlas integrated in CAT12.

*2.3.4. Eigenvariate extraction*

Significant hippocampal clusters from the CN > AD and MCI > AD contrasts were selected for region-of-interest (ROI) analysis. From each peak, eigenvariates were extracted using SPM's volume of interest (VOI) tool, retaining the respective contrast while regressing out age and TIV. These adjusted eigenvariates summarize the predominant variance in the cluster for each subject. Pairwise group differences on the eigenvariates were assessed with two-sample t-tests, and effect sizes were quantified using Cohen's d which is defined as the difference in group means divided by the pooled standard deviation [11].

$$Cohen's\ d\ =\ \frac{\overline{x}_1 - \overline{x}_2}{\sqrt{\frac{s_1^2 + s_2^2}{2}}}$$

Where,

$\overline{x}_i$ = Mean of Group 1 and 2

$\sqrt{\frac{s_1^2 + s_2^2}{2}}$ = Standard Deviation of Group 1 and 2

## 3. RESULTS

### 3.1. WHOLE-BRAIN VBM FINDING

In the **CN > AD** contrast, cognitively normal participants showed significantly greater volume than Alzheimer's disease patients in the left hippocampus (peak MNI coordinate: [−2,4,−2], cluster extent = 52,045 voxels, peak T = 11.00, cluster-level FWE-corrected p<0.001). Adjusted eigenvariates extracted from this cluster revealed a large group difference (CN mean ± SE = 0.3947±0.0044, AD mean ± SE = 0.3106±0.0075, t=9.58, p<0.0001, Cohen's d=2.03). SE represents standard deviation.

In the **MCI > AD** contrast, mild cognitive impairment participants had higher hippocampal volume than AD patients in the left hippocampus (peak MNI coordinate: [−12,0,−9], cluster extent = 26,799 voxels, peak T = 9.15, cluster-level FWE-corrected p<0.001). The eigenvariate comparison, adjusted for age and TIV, confirmed a robust effect (MCI mean ± SE = 0.3886±0.0040, AD mean ± SE = 0.3187±0.0075, t=7.69, p < 0.0001, Cohen's d=1.61).

The **CN > MCI** contrast did not yield any suprathreshold clusters in the hippocampus or elsewhere at voxelwise *p*<0.001, cluster-level FWE-corrected *p*<0.05, suggesting a more subtle volume difference than observed in the MCI > AD or CN > AD comparisons.

As visualized in Figure 2 and Figure 4, CN and MCI participants showed significantly higher gray matter volume than AD subjects in widespread regions including the bilateral hippocampus, medial temporal lobe, and frontal cortex. These differences were obtained from the *CN > AD and MCI > AD* contrast with log false discovery rate (FDR) correction.

**Table 3A.** Hippocampal imaging results for CN > AD and MCI > AD contrasts. Peak MNI coordinates are in MNI space; cluster extent is in voxels; statistics are thresholded voxelwise at p<0.001 with cluster-level FWE correction.

| Contrast | Region | Peak MNI (x,y,z) | Cluster size (voxels) | Peak T | Cluster-level FWE p |
|---|---|---|---|---|---|
| CN>AD | Hippocampus (left) | [-2, 4, -2] | 52045 | 11.00 | <0.001 |
| MCI>AD | Hippocampus (left) | [-12 0 -9] | 26799 | 9.15 | <0.001 |

**Table 3B.** Eigenvariate-based hippocampal volume comparisons (adjusted for age and total intracranial volume). Group means ± SE, t-statistics, p-values, and effect sizes (Cohen's d) are shown.

| Contrast | Group 1 mean ± SE | Group 2 mean ± SE | t | p | Cohen's d |
|---|---|---|---|---|---|
| CN>AD | 0.3947 ± 0.0044 | 0.3106 ± 0.0075 | 9.575 | <0.0001 | 2.03 |
| MCI>AD | 0.3886 ± 0.0040 | 0.3187 ± 0.0075 | 7.69 | <0.0001 | 1.61 |

### 3.2. ROI AND CLUSTER-ANALYSIS

Thresholded statistical maps were used to identify significant clusters, which were saved as binary masks. A prominent left hippocampal cluster from the CN > AD and MCI > AD contrasts was selected for region-of-interest (ROI) analysis. From each peak, eigenvariates adjusted for age and total intracranial volume were extracted using SPM's VOI tool. These eigenvariates summarize the dominant variance within the hippocampal cluster for each subject. Group comparisons on the adjusted eigenvariates showed a graded pattern of hippocampal volume: highest in CN, intermediate in MCI, and lowest in AD. The absence of a suprathreshold hippocampal cluster in the CN > MCI contrast further supports this progressive trajectory. Figure 2 and Figure 4 illustrate the statistical maps with contrast schematics, design matrices, and the extracted eigenvariate time series for the significant hippocampal clusters. Figure 3 and Figure 5 illustrate the anatomical location of the ROIs.

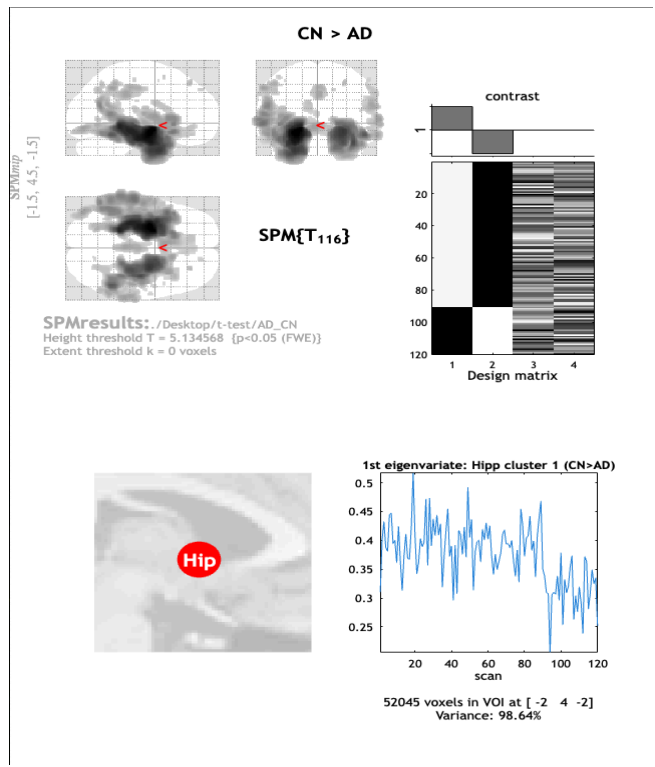

**Fig.2.** Gray matter volume differences for CN >AD. Statistical maps (top) show the clusters and corresponding design matrices and contrasts are displayed. Bottom shows the eigenvariates from each cluster after regressing out age and TIV.

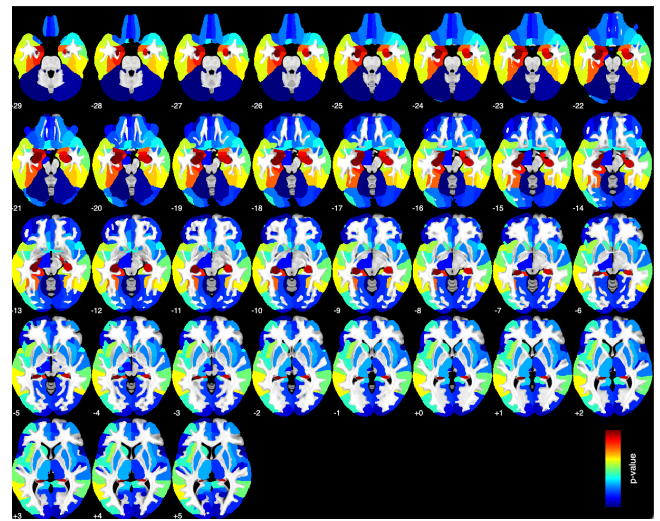

**Fig. 3.** Neuromorphometrics atlas for CN >AD

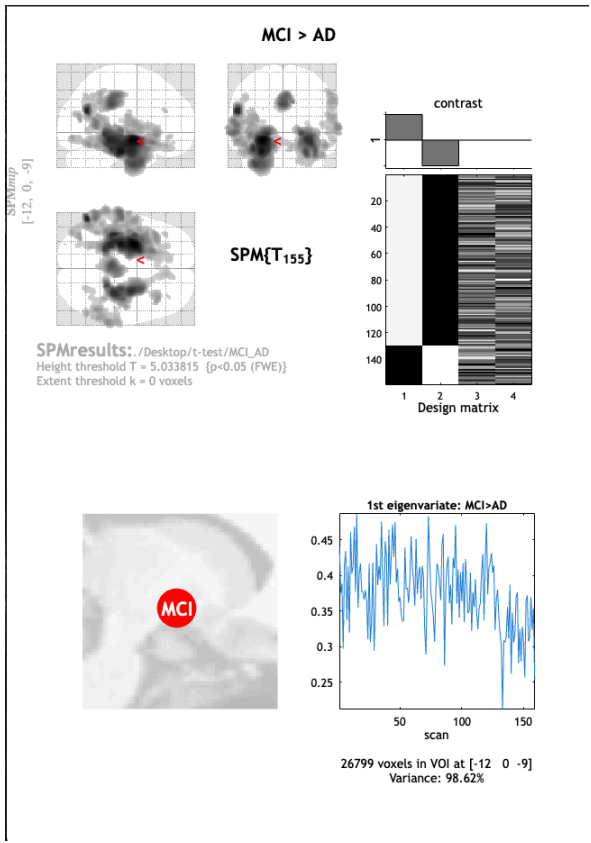

**Fig. 4.** Gray matter volume differences for MCI >AD. Statistical maps (top) show the clusters and corresponding design matrices and contrasts are displayed. Bottom shows the eigenvariates from each cluster after regressing out age and TIV.

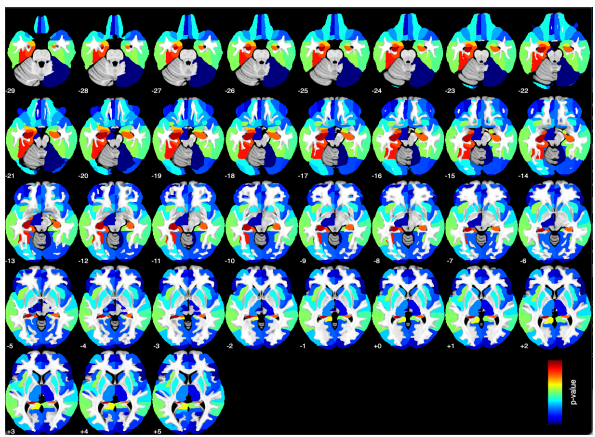

**Fig. 5.** Neuromorphometrics atlas for MCI >AD

### 3.3. PREDICTIVE ANALYSIS OF MCI-TO-AD CONVERSION

The predictive utility of hippocampal volume for identifying patients with mild cognitive impairment (MCI) at risk of progressing to Alzheimer's disease was assessed by evaluating three prediction models using 5-fold cross-validated logistic regression. Out of the 129 MCI subjects, 9 (7.0%) converted to AD within 24 months while 120 remained cognitively stable, consistent with conversion rates observed in ADNI cohorts with short follow-up [18]. Three models were compared:

(1) Clinical features only (age, sex, education, Mini-Mental State Examination (MMSE), APOE4 carrier status),

(2) Hippocampal eigenvariate only, and

(3) Combined clinical and imaging features.

The model that used only hippocampal volume showed moderate ability to distinguish between those who would develop AD and those who would not (AUC = 0.662 ± 0.214), and it performed better than the model with only clinical features (AUC = 0.505 ± 0.250). However, combining both types of data did not improve prediction (AUC = 0.546 ± 0.217; all pairwise $p > 0.27$). In the hippocampal-only model, having a smaller baseline hippocampal volume was linked to a higher risk of developing AD ($\beta$ = -0.551, OR = 0.576). This model had 70% sensitivity but only 9.7% precision, attributable to the low base rate of conversion (7%). Overall, these results demonstrate that while hippocampal atrophy provides prognostic information [19, 20], its predictive utility remains modest, and the limited number of conversion events constrains its applicability for individual risk stratification.

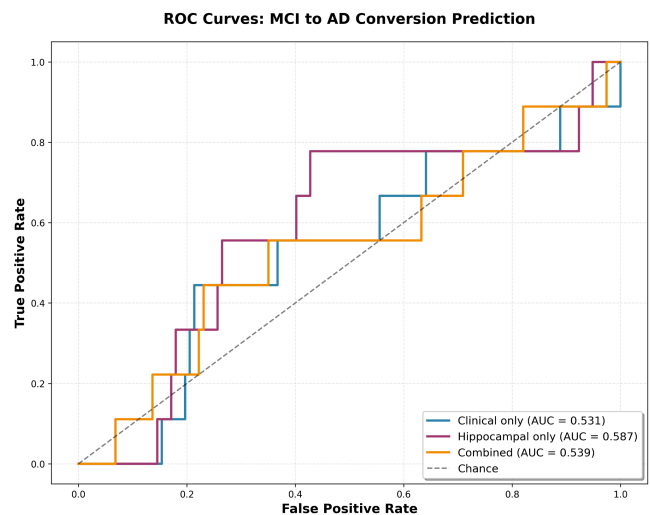

**Fig. 6.** ROC curves for MCI-to-AD conversion prediction.

Three prediction models were evaluated using 5-fold cross-validation: clinical features only (blue), eigenvariate only (purple), and combined clinical and imaging features (orange). Hippocampal volume showed the highest discriminative ability (AUC = 0.587), though performance remained modest overall. The dashed line represents chance performance (AUC= 0.50).

**Table 4a.** Predictive performance for MCI-to-AD conversion

| Model | AUC | Accuracy | Precision | Recall | F1 Score |
|---|---|---|---|---|---|
| Clinical only | 0.531 | 0.650 | 0.075 | 0.40 | 0.126 |
| Hippocampal only | 0.587 | 0.587 | 0.097 | 0.70 | 0.168 |
| Combined | 0.539 | 0.674 | 0.071 | 0.40 | 0.120 |

### 3.4. APOE4 Stratification Analysis

The cohort was stratified by APOE4 carrier status to determine whether the APOE4 genotype modulates hippocampal atrophy patterns. The frequency of APOE4 carriers increased with disease severity (CN: 32.2%, MCI: 33.3%, AD: 46.7%), which aligns with established genetic risk patterns for Alzheimer's disease [33].

Independent samples t-test were conducted within each diagnostic group to compare hippocampal volumes between APOE4 carriers and non-carriers. No significant differences were observed in any group: CN (t = -0.32, p = 0.748, Cohen's d = -0.07), MCI (t = 0.13, p = 0.896, Cohen's d = 0.02), and AD ( t = 0.04, p = 0.969, Cohen's d = 0.01). Effect sizes were negligible across all comparisons.

A two-way ANOVA was performed with diagnosis and APOE4 status as independent variables and hippocampal volume as the dependent variable. The analysis revealed a significant main effect of diagnosis (F(2,243) = 44.25, p < 0.001), consistent with progressive hippocampal atrophy across the disease spectrum. No significant main effect of APOE4 status (F(1,243) = 0.005, p = 0.943) or diagnosis by APOE4 interaction (F(2,243) = 0.055, p = 0.947) was detected, which suggests that APOE4 carrier status did not significantly influence hippocampal volume in this cross-sectional analysis.

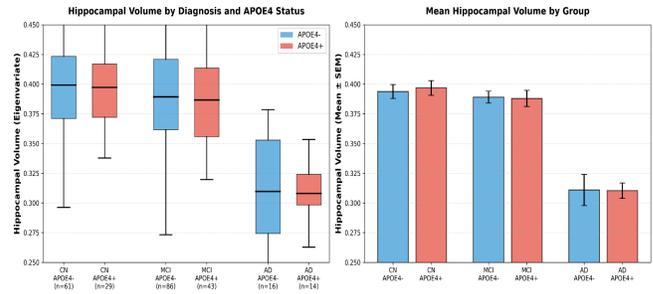

**Fig. 7.** Hippocampal volume stratified by diagnosis and APOE4 carrier status

On the left, box plots show the distributions of hippocampal eigenvariate values for APOE4 non-carriers (blue) and carriers (red) in each diagnostic group, with sample sizes in parentheses. On the right, mean hippocampal volume is shown with standard error bars. There were no significant differences between APOE4 groups in any diagnosis (all p > 0.74).

**Table 4b.** APOE4 stratification results

| Group | APOE4- | APOE4+ | Mean Difference | t-statistic | p-value | Cohen's d |
|---|---|---|---|---|---|---|
| CN | n=61: 0.394 ± 0.046 | n=29: 0.397 ± 0.034 | -0.003 | -0.32 | 0.748 | -0.07 |
| MCI | n=86: 0.389 ± 0.047 | n=43: 0.388 ± 0.044 | +0.001 | 0.13 | 0.896 | 0.02 |
| AD | n=16: 0.311 ± 0.053 | n=14: 0.310 ± 0.024 | +0.001 | 0.04 | 0.969 | 0.01 |

*NOTE: Values are mean ± standard deviation. Positive mean differences indicate higher volume in APOE4 non-carriers.*

## 4. DISCUSSION AND CONCLUSION

This VBM analysis supports previous findings that the hippocampus and entorhinal atrophy as key markers of progressive neurodegeneration in the AD spectrum [4,6,16,17]. The large effect sizes observed in the CN > AD and MCI > AD contrasts (Cohen's d = 2.03 and 1.61, respectively) show that medial temporal structure volume clearly distinguishes disease severity, even after adjusting for age and intracranial volume. The lack of a significant

CN > MCI hippocampal effect suggests a more gradual decline from CN to MCI, with the most noticeable atrophy occurring as patients progress to AD.

The result that combining clinical and imaging models did not do better than imaging alone is important to note. This could be because hippocampal volume and cognitive scores both measure disease severity, so they overlap, or because the model was not complex enough to capture more complicated relationships. In the future, using changes in atrophy over time, adding more types of biomarkers (like amyloid PET and tau PET), and applying advanced machine learning methods may help improve prediction.

Although the model's performance was only modest, these results show that VBM-based hippocampal measurements provide useful prognostic information that routine clinical assessments do not capture. With more participants and longer follow-up, these methods could help make risk assessments more specific for clinical trials or targeted treatments.

### 4.1. APOE4 and Structural Neurodegeneration

Contrary to some prior reports [21,22], no significant modulation of hippocampal volume by APOE4 carrier status (all $p > 0.74$) was observed. There are a few possible reasons for this result.

First, APOE4 mainly affects amyloid-β buildup and removal [23,24], while changes in brain structure happen later through steps like tau pathology, neuroinflammation, and synaptic dysfunction [25,26]. Measuring brain volume at a single time point may not be sensitive enough to detect the indirect effects of APOE4.

Second, studies that track changes in hippocampal atrophy over time have more often found effects of APOE4 [27,28,29]. This suggests that APOE4 carrier status affects how quickly neurodegeneration happens, rather than the total volume at one point in time. Because this study only looked at one time point, it could not examine changes over time.

Third, the Alzheimer's disease group was small (n=30), and the number of APOE4 carriers was moderate (47%), which made it harder to detect interaction effects. Usually, finding these effects needs larger sample sizes than finding main effects [30].

The significant differences between diagnostic groups ($F = 44.25$, $p < 0.001$), along with the lack of APOE4 effects, suggest that hippocampal atrophy shows the overall disease burden, no matter the genetic risk factors. This finding supports the idea that APOE4 speeds up the onset and progression of the disease [31,32], but once clinical dementia appears, the structural changes in the brain are similar across different genotypes. This "convergence hypothesis" implies that APOE4 determines when one reaches a given atrophy threshold, but not the threshold itself.

### 5. LIMITATIONS

This study has some limitations. Because it uses a cross-sectional design, it was not possible to track changes in atrophy or APOE4 effects on decline rates within individuals, which might be more sensitive than just looking at baseline volumes. Also, excluding subjects with corrupted or non-3D files may have caused some sampling bias.

There are also some specific limitations in the predictive and stratification analyses. The low rate of MCI-to-AD conversion (7%, n=9/129) is due to the short follow-up period, which reduces the statistical power for prediction. Larger groups with longer follow-up would help make the results stronger. Biomarker confirmation (amyloid or tau PET) was not available to distinguish prodromal AD from other MCI etiologies, which may have weakened observed associations. The predictive performance was modest (AUC about 0.66), showing that hippocampal volume alone is not enough for individual predictions. This suggests that combining clinical, genetic, and imaging biomarkers would be more effective. Finally, the small number of AD cases (n=30) made it harder to detect effects related to APOE4.

### 6. FUTURE WORKS

Future research will extend this cross-sectional framework in several directions. First, longitudinal modeling will be incorporated to characterize within-subject evolution and improve the temporal resolution of atrophy progression, with particular attention to APOE4-stratified atrophy rates. Second, predictive models that combine hippocampal eigenvariates with other imaging and clinical features will be developed and validated to forecast conversion from mild cognitive impairment (MCI) to Alzheimer's disease (AD). Third, multimodal data, including Positron Emission Tomography (PET) biomarkers, diffusion metrics, and cognitive scores, will be integrated to enhance specificity and sensitivity. Fourth, small volume correction and regionally informed priors will be applied to improve detection of early-stage changes. The analyses will be expanded to include diverse and larger cohorts to assess generalizability and population heterogeneity.

### 7. REFERENCES

1. Mayo Clinic Staff, "Alzheimer's disease - Symptoms and causes," Mayo clinic, Jul. 2024. https://www.mayoclinic.org/diseases-conditions/alzheimers-disease/symptoms-causes/syc-20350447
2. J. L. Whitwell, "Voxel-Based Morphometry: An Automated Technique for Assessing Structural Changes in the Brain," *J. Neurosci.*, vol. 29, no. 31, pp. 9661–9664, Aug. 2009, doi:10.1523/JNEUROSCI.2160-09.2009.
3. Jack, C. R., et al. "The Alzheimer's Disease Neuroimaging Initiative (ADNI): MRI methods." Journal of Magnetic Resonance Imaging 27.4 (2008): 685-691.

4. Rathore, S., et al. "A review on neuroimaging-based classification studies and associated feature extraction methods for Alzheimer's disease and its prodromal stages." NeuroImage 155 (2017): 530-548.
5. Frisoni, G. B., et al. "The clinical use of structural MRI in Alzheimer disease." Nature Reviews Neurology 6.2 (2010): 67-77.
6. Ribeiro, L. G., & Busatto Filho, G. (2016). Voxel-based morphometry in Alzheimer's disease and mild cognitive impairment: Systematic review of studies addressing the frontal lobe. *Dementia & Neuropsychologia*, 10(2), 104–112.
7. Gaser C, Dahnke R, Thompson PM, Kurth F, Luders E, Alzheimer"s Disease Neuroimaging Initiative (2024). CAT: a computational anatomy toolbox for the analysis of structural MRI data. GigaScience 13, giae049.
8. Friston, K. J., Ashburner, J., Kiebel, S., Nichols, T., & Penny, W. (2007). *Statistical parametric mapping: The analysis of functional brain images* [Software]. Wellcome Centre for Human Neuroimaging, University College London. https://fil.ion.ucl.ac.uk/spm/
9. Ashburner, J., & Friston, K. J. (2000). Voxel-based morphometry—The methods. *NeuroImage*, 11(6 Pt 1), 805–821
10. Cohen, J. (1988). Statistical Power Analysis for the Behavioral Sciences (2nd ed.). Hillsdale, NJ: Lawrence Erlbaum Associates
11. Nelder, J. A., & Wedderburn, R. W. M. (1972). Generalized Linear Models. Journal of the Royal Statistical Society, Series A (General), 135(3), 370–384.
12. Alzheimer's Association. (2024). 2024 Alzheimer's disease facts and figures. *Alzheimer's & Dementia*, 20(5), 3708-3821.
13. Braak, H., & Braak, E. (1991). Neuropathological stageing of Alzheimer-related changes. *Acta Neuropathologica*, 82(4), 239-259.
14. Scheltens, P., De Strooper, B., Kivipelto, M., Holstege, H., Chételat, G., Teunissen, C. E., ... & Blennow, K. (2021). Alzheimer's disease. *The Lancet*, 397(10284), 1577-1590.
15. Good, C. D., Johnsrude, I. S., Ashburner, J., Henson, R. N., Friston, K. J., & Frackowiak, R. S. (2001). A voxel-based morphometric study of ageing in 465 normal adult human brains. *NeuroImage*, 14(1), 21-36.
16. Apostolova, L. G., Dinov, I. D., Dutton, R. A., Hayashi, K. M., Toga, A. W., Cummings, J. L., & Thompson, P. M. (2006). 3D comparison of hippocampal atrophy in amnestic mild cognitive impairment and Alzheimer's disease. *Brain*, 129(11), 2867-2873.
17. Convit, A., De Asis, J., De Leon, M. J., Tarshish, C. Y., De Santi, S., & Rusinek, H. (2000). Atrophy of the medial occipitotemporal, inferior, and middle temporal gyri in non-demented elderly predict decline to Alzheimer's disease. *Neurobiology of Aging*, 21(1), 19-26.
18. Petersen, R. C., Aisen, P. S., Beckett, L. A., Donohue, M. C., Gamst, A. C., Harvey, D. J., ... & Weiner, M. W. (2010). Alzheimer's Disease Neuroimaging Initiative (ADNI): clinical characterization. *Neurology*, 74(3), 201-209.
19. Devanand, D. P., Pradhaban, G., Liu, X., Khandji, A., De Santi, S., Segal, S., ... & Pelton, G. H. (2007). Hippocampal and entorhinal atrophy in mild cognitive impairment: prediction of Alzheimer disease. *Neurology*, 68(11), 828-836.
20. Trzepacz, P. T., Yu, P., Sun, J., Schuh, K., Case, M., Witte, M. M., ... & Degenhardt, E. (2014). Comparison of neuroimaging modalities for the prediction of conversion from mild cognitive impairment to Alzheimer's dementia. *Neurobiology of Aging*, 35(1), 143-151.
21. Morra, J. H., Tu, Z., Apostolova, L. G., Green, A. E., Avedissian, C., Madsen, S. K., ... & Thompson, P. M. (2009). Automated mapping of hippocampal atrophy in 1-year repeat MRI data from 490 subjects with Alzheimer's disease, mild cognitive impairment, and elderly controls. *NeuroImage*, 45(1), S3-S15.
22. Crivello, F., Lemaître, H., Dufouil, C., Grassiot, B., Delcroix, N., Tzourio-Mazoyer, N., ... & Mazoyer, B. (2010). Effects of ApoE-ε4 allele load and age on the rates of grey matter and hippocampal volumes loss in a longitudinal cohort of 1186 healthy elderly persons. *NeuroImage*, 53(3), 1064-1069.
23. Liu, C. C., Kanekiyo, T., Xu, H., & Bu, G. (2013). Apolipoprotein E and Alzheimer disease: risk, mechanisms and therapy. *Nature Reviews Neurology*, 9(2), 106-118.
24. Yamazaki, Y., Zhao, N., Caulfield, T. R., Liu, C. C., & Bu, G. (2019). Apolipoprotein E and Alzheimer disease: pathobiology and targeting strategies. *Nature Reviews Neurology*, 15(9), 501-518.
25. Jack Jr, C. R., Knopman, D. S., Jagust, W. J., Petersen, R. C., Weiner, M. W., Aisen, P. S., ... & Trojanowski, J. Q. (2013). Tracking pathophysiological processes in Alzheimer's disease: an updated hypothetical model of dynamic biomarkers. *The Lancet Neurology*, 12(2), 207-216.
26. Shi, Y., Yamada, K., Liddelow, S. A., Smith, S. T., Zhao, L., Luo, W., ... & Holtzman, D. M. (2017). ApoE4 markedly exacerbates tau-mediated neurodegeneration in a mouse model of tauopathy. *Nature*, 549(7673), 523-527.
27. Jack Jr, C. R., Wiste, H. J., Weigand, S. D., Knopman, D. S., Vemuri, P., Mielke, M. M., ... & Petersen, R. C. (2015). Age, sex, and APOE ε4 effects on memory, brain structure, and β-amyloid across the adult life span. *JAMA Neurology*, 72(5), 511-519.
28. Manning, E. N., Barnes, J., Cash, D. M., Bartlett, J. W., Leung, K. K., Ourselin, S., & Fox, N. C. (2014). APOE ε4 is associated with disproportionate progressive hippocampal atrophy in AD. *PloS One*, 9(5), e97608.
29. Fjell, A. M., McEvoy, L., Holland, D., Dale, A. M., Walhovd, K. B., & Alzheimer's Disease Neuroimaging Initiative. (2014). What is normal in normal aging? Effects of aging, amyloid and Alzheimer's disease on the cerebral cortex and the hippocampus. *Progress in Neurobiology*, 117, 20-40.
30. Leon, A. C., & Heo, M. (2009). Sample sizes required to detect interactions between two binary fixed-effects in a mixed-effects linear regression model. *Computational Statistics & Data Analysis*, 53(3), 603-608.
31. Qiu, C., Kivipelto, M., Agüero-Torres, H., Winblad, B., & Fratiglioni, L. (2004). Risk and protective effects of the APOE gene towards Alzheimer's disease in the Kungsholmen project: variation by age and sex. *Journal of Neurology, Neurosurgery & Psychiatry*, 75(6), 828-833.
32. Verghese, P. B., Castellano, J. M., & Holtzman, D. M. (2011). Apolipoprotein E in Alzheimer's disease and other neurological disorders. *The Lancet Neurology*, 10(3), 241-252.
33. Farrer, L. A., Cupples, L. A., Haines, J. L., Hyman, B., Kukull, W. A., Mayeux, R., ... & Van Duijn, C. M. (1997). Effects of age, sex, and ethnicity on the association between apolipoprotein E genotype and Alzheimer disease. JAMA, 278(16), 1349-1356.